\documentclass[10pt,twocolumn]{article}
\usepackage{cite}
\usepackage{amsmath,amssymb,amsfonts}
\usepackage{graphicx}
\usepackage{textcomp}
\usepackage{multirow}
\usepackage{array}
\usepackage[margin=0.75in]{geometry}
\usepackage{subcaption}
\usepackage{times}
\def\BibTeX{{\rm B\kern-.05em{\sc i\kern-.025em b}\kern-.08em
    T\kern-.1667em\lower.7ex\hbox{E}\kern-.125emX}}

\begin{document}
\title{NanoSleep: A Parameter-Efficient Hybrid Temporal Convolutional Network for Single-Channel Sleep Stage Classification}

\author{%
S M Asif Hossain and Shruti Kshirsagar\\[0.4em]
\small\textit{School of Computing, Wichita State University, Kansas, USA}
}
\date{}

\maketitle

\begin{abstract}
Sleep stage classification from single-channel electroencephalography (EEG) is essential for wearable and home-based sleep monitoring. However, many deep learning models achieve high accuracy at the cost of large model sizes, which limits their deployment on resource-constrained devices. In this work, we present NanoSleep, a compact hybrid temporal convolutional network for automatic sleep stage classification. NanoSleep combines a learnable Sinc-convolutional front end, a dual-branch feature extractor that fuses multi-scale temporal and spectral representations, a gated dilated temporal convolutional backbone with channel recalibration, and a conditional random field for sequence-level decoding. We further employ a weighted calibrated focal loss to address class imbalance. We evaluate NanoSleep on the Sleep-EDF and Sleep-EDF-Expanded datasets using subject-wise cross-validation. The proposed model consistently outperforms six representative baseline methods, and an ablation study confirms the contribution of each major component. These results demonstrate that NanoSleep provides an effective balance between accuracy and efficiency, making it well suited for wearable devices, home-based sleep monitoring, and resource-constrained clinical applications.
\end{abstract}


\section{Introduction}
\label{sec:introduction}

Sleep is a fundamental physiological process that supports memory consolidation, immune regulation, and the restoration of cognitive and metabolic functions \cite{luyster2012sleep}. Poor or insufficient sleep is associated with cardiovascular disease, cognitive decline, and several neurological and psychiatric disorders \cite{wulff2010sleep, bertisch2018insomnia}. Accurate sleep assessment is therefore essential for diagnosing and managing sleep-related disorders. The first step in this process is sleep stage classification, which divides an overnight recording into a sequence of discrete sleep stages. Polysomnography (PSG) is the clinical standard for sleep assessment. It simultaneously records the electroencephalogram (EEG), electrooculogram (EOG), electromyogram (EMG), and other physiological signals throughout the night. According to the American Academy of Sleep Medicine (AASM) guidelines, the recording is divided into consecutive 30-second epochs. Trained sleep technologists assign each epoch to one of five stages: wakefulness (W), non-rapid-eye-movement sleep (N1, N2, and N3), or rapid-eye-movement sleep (REM) \cite{berry2017aasm}. Although this procedure provides reliable clinical assessment, it is labor-intensive, time-consuming, and subject to inter-scorer variability. Agreement between experienced scorers often falls below 90\% \cite{rosenberg2013american}. Moreover, PSG requires specialized laboratories, expensive equipment, and multiple electrodes that may disturb natural sleep. These limitations have motivated the development of automatic sleep staging systems that operate with fewer physiological signals. Among the available physiological signals, EEG provides the most discriminative information for sleep staging because characteristic waveforms such as sleep spindles, K-complexes, and slow waves are directly reflected in brain activity \cite{phan2022automatic}. Single-channel EEG is particularly attractive for wearable and home-based monitoring because it reduces hardware complexity and improves user comfort while preserving most stage-relevant information \cite{imtiaz2021systematic, mikkelsen2019accurate}. As a result, automatic sleep stage classification using single-channel EEG has become an active area of research.
Recent deep learning methods have substantially improved automatic sleep staging by learning discriminative representations directly from raw EEG signals. Convolutional neural networks effectively capture local temporal patterns, recurrent neural networks model sequential sleep dynamics, and attention-based architectures further improve contextual modeling. Despite these advances, several important challenges remain.

The first challenge is computational efficiency. High-performing models such as SleepEEGNet \cite{mousavi2019sleepeegnet} and XSleepNet \cite{phan2021xsleepnet} contain millions of trainable parameters. Such models are difficult to deploy on wearable devices and resource-constrained clinical hardware because of their memory, computational, and energy requirements. Although lightweight alternatives have been proposed \cite{supratak2020tinysleepnet, fiorillo2021deepsleepnetlite}, they often sacrifice classification accuracy to reduce model size.

The second challenge is class imbalance. Sleep datasets contain substantially fewer N1 epochs than the other sleep stages. Consequently, many deep learning models perform poorly on this minority class \cite{zhou2020automatic, sun2019two}. The third challenge is sequence modeling. Many methods classify each epoch independently and therefore ignore the physiological transition rules that govern normal sleep progression \cite{kuo2020automatic}. As a result, these models may produce isolated predictions that are inconsistent with biological sleep patterns.

To address these challenges, we propose NanoSleep, a compact hybrid network for automatic sleep stage classification from single-channel EEG. NanoSleep combines learned multi-scale temporal representations with interpretable spectral features to improve feature learning while maintaining a small model size. The architecture integrates a learnable Sinc-convolutional front end, a dual-branch feature extractor, a gated dilated temporal convolutional backbone with channel recalibration, and a conditional random field (CRF) decoder for sequence-level prediction. We further employ a weighted calibrated focal loss to address severe class imbalance during training.

The main contributions of this work are summarized as follows.

\begin{enumerate}
\item We propose NanoSleep, a compact hybrid architecture that combines learned multi-scale temporal representations with interpretable spectral features. The proposed model employs a gated dilated temporal convolutional backbone and a CRF decoder to achieve high classification accuracy with only 0.35~M trainable parameters.

\item We develop an efficient training framework that integrates learnable signal preprocessing, imbalance-aware optimization, and sequence-level decoding. An ablation study quantifies the contribution of each component and validates the proposed architecture.

\item We evaluate NanoSleep on the Sleep-EDF and Sleep-EDF-Expanded datasets using subject-wise cross-validation. The proposed model consistently outperforms representative state-of-the-art methods while requiring substantially fewer parameters.
\end{enumerate}

The remainder of this paper is organized as follows. Section~\ref{sec:related} reviews related work. Section~\ref{sec:method} presents the proposed NanoSleep framework. Section~\ref{sec:setup} describes the experimental setup. Section~\ref{sec:results} reports and discusses the experimental results. Section~\ref{sec:limitations} outlines the study limitations and future research directions. Finally, Section~\ref{sec:conclusion} concludes the paper.

\section{Related Works}
\label{sec:related}

In this section, we discuss the related work in the sleep stage classification domain.
\subsection{Conventional Feature-Based Classifiers}
Early automatic sleep staging methods relied on handcrafted EEG features combined with conventional machine learning classifiers. Researchers extracted time-domain statistics, modulation spectrogram, spectral band powers, entropy measures, and nonlinear complexity features to characterize sleep stages \cite{hassan2016decision, lajnef2015learning, tallal2026modulation}. Memar and Faradji combined spectral features with a random forest classifier and reported promising performance \cite{memar2018novel}. Hassan and Bhuiyan employed ensemble empirical mode decomposition with boosting classifiers for single-channel EEG sleep staging \cite{hassan2017automated}. Jiang et al. incorporated temporal information through a hidden Markov model \cite{jiang2019robust}, while Zhou et al. proposed a stacked ensemble with a class-balancing strategy to improve N1 recognition \cite{zhou2020automatic}. Although these methods are computationally efficient and interpretable, their performance depends heavily on manually designed features and often generalizes poorly across different datasets and recording conditions. These limitations motivated the development of deep learning methods that learn discriminative representations directly from EEG signals.
% Conventional approaches to sleep staging combine handcrafted features with a statistical classifier. A wide range of descriptors has been explored, including time-domain statistics, spectral band powers, entropy measures, and nonlinear complexity indices \cite{hassan2016decision, lajnef2015learning}. Memar and Faradji extracted spectral features from sub-bands of the EEG and reported strong performance with a random forest classifier \cite{memar2018novel}. Hassan and Bhuiyan investigated ensemble empirical mode decomposition together with boosting classifiers and demonstrated robust staging from single-channel EEG \cite{hassan2017automated}. Jiang et al. coupled a multimodal decomposition with a hidden Markov model that refined the per-epoch predictions using sequential context \cite{jiang2019robust}, and Zhou et al. proposed a two-layer stacked ensemble of a random forest and a gradient boosting model, together with a class-balance strategy that improved the recognition of the N1 stage \cite{zhou2020automatic}. While these methods are interpretable and computationally light, their accuracy is bounded by the quality of the manually designed features, and they often transfer poorly across datasets and acquisition hardware. This limitation motivated the adoption of representation-learning methods that discover stage-relevant features directly from the signal.

\subsection{Deep Learning on Single-Channel EEG}

Deep learning has become the dominant approach for automatic sleep stage classification. Tsinalis et al. introduced stacked sparse autoencoders for learning EEG representations from time-frequency images \cite{tsinalis2016automatic}. Sors et al. later demonstrated that convolutional neural networks can classify sleep stages directly from raw single-channel EEG without handcrafted features \cite{sors2018convolutional}. Supratak et al. proposed DeepSleepNet, which combines multi-scale convolutional feature extraction with bidirectional long short-term memory networks for temporal modeling \cite{supratak2017deepsleepnet}. Sandhu et al. \cite{sandhuexploring} demonstrated the effectiveness of single-channel EEG for accurate, interpretable, and practical automated sleep staging. Subsequent studies further improved performance using deeper architectures and larger training datasets \cite{biswal2018expert, stephansen2018neural}. More recently, Yang et al. introduced BIOT, a transformer-based backbone for large-scale biosignal representation learning that improves feature transfer across heterogeneous physiological datasets \cite{yang2024biot}. These studies demonstrate the effectiveness of learned representations but generally require large models with high computational and memory costs.
% Deep learning has become the dominant paradigm for single-channel EEG sleep staging. Tsinalis et al. applied stacked sparse autoencoders to time-frequency representations of the EEG \cite{tsinalis2016automatic}, and Sors et al. showed that a convolutional neural network can score sleep stages from raw single-channel EEG without any handcrafted features \cite{sors2018convolutional}. Supratak et al. proposed DeepSleepNet, a two-stream convolutional feature extractor followed by bidirectional long short-term memory layers that model the temporal structure of a night of sleep \cite{supratak2017deepsleepnet}. The DeepSleepNet family has been influential and remains a common baseline. Biswal et al. subsequently demonstrated that a deep network trained on a very large corpus can approach expert-level scoring performance \cite{biswal2018expert}, and Stephansen et al. showed that a deep model trained on a broad polysomnography cohort enables efficient diagnosis of narcolepsy \cite{stephansen2018neural}. These works collectively established that learned representations can outperform handcrafted features, but the resulting networks are large, and their training and inference footprints exceed the resources of typical wearable and point-of-care platforms.

\subsection{Sequence and Attention Architectures}
Sleep stages exhibit strong temporal dependencies, motivating sequence-based learning methods. Phan et al. proposed SeqSleepNet, a hierarchical recurrent network for sequence-to-sequence sleep staging \cite{phan2019seqsleepnet}. Mousavi et al. introduced SleepEEGNet, an attention-based encoder-decoder architecture that improves classification performance at the cost of a large parameter count \cite{mousavi2019sleepeegnet}. Eldele et al. developed AttnSleep by combining convolutional feature extraction with multi-head attention \cite{eldele2021attention}. Phan et al. later proposed XSleepNet and SleepTransformer, which integrate multiple signal representations and transformer-based sequence modeling to achieve state-of-the-art performance \cite{phan2021xsleepnet, phan2022sleeptransformer}. Graph-based approaches, including GraphSleepNet and SalientSleepNet, have also been proposed to model spatial-temporal relationships and salient EEG patterns \cite{jia2020graphsleepnet, jia2021salientsleepnet}.

Recent studies have further improved model generalization through transfer learning and domain adaptation. Hossain and Kshirsagar proposed a demographic-aware transfer learning framework for cross-cohort sleep staging \cite{hossain2026demographic}. Eldele et al. introduced ADAST, an attentive domain adaptation framework based on iterative self-training \cite{eldele2023adast}. Tallal et al. proposed STDA-Net for unsupervised domain adaptation across sleep datasets \cite{tallal2026stda}. These studies demonstrate the importance of robust feature learning for clinical deployment. However, most sequence and attention models achieve higher accuracy by substantially increasing model complexity.

\subsection{Data Augmentation Strategies}
Data augmentation has been widely adopted across various domains to enhance the generalization capability and robustness of models trained on speech \cite{kshirsagar2022cross}, \cite{kshirsagar2023task}, image \cite{mouradi2026robust}, \cite{parupatitowards}, and physiological signal data \cite{sun2019two, mainsmachine}. Data augmentation has been used to increase training diversity and reduce the severe imbalance among sleep stages. Sun et al. introduced an oversampling-based pretraining procedure that applies temporal shifts and additive white noise to minority-stage EEG epochs \cite{sun2019two}. Fan et al. systematically evaluated repeated sampling, morphological transformations, signal segmentation and recombination, cross-dataset transfer, and generative adversarial network-based synthesis on the MASS and Sleep-EDF datasets \cite{fan2020eeg}. Lee et al. proposed spectral band blending, in which selected frequency bands are exchanged between EEG signals to generate new samples while preserving stage-related spectral information \cite{lee2021spectral}. Khalili and Mohammadzadeh Asl also incorporated augmented sequence samples into a temporal convolutional framework to improve the training of raw-EEG classifiers \cite{khalili2021automatic}.

Generative and noise-based augmentation methods have received increasing attention \cite{kshirsagar2022affective}, \cite{kshirsagar2022quality}. Ling et al. developed an improved deep convolutional generative adversarial network that generates continuous-wavelet time-frequency maps for minority sleep stages \cite{ling2022staging}. Huang et al. applied Gaussian noise augmentation to underrepresented polysomnography segments to improve class balance and N1 recognition \cite{huang2023sleep}. Rommel et al. compared 13 EEG augmentation transformations and showed that their effectiveness depends on the task, dataset, and training regime \cite{rommel2022augmentation}. These studies demonstrate that data augmentation can improve robustness and minority-stage recognition. However, the selected transformations must preserve physiologically meaningful EEG morphology and temporal structure.

\subsection{Efficient and Compact Models}
Several studies have focused on reducing the computational cost of deep sleep staging models. TinySleepNet substantially reduces the parameter count of DeepSleepNet while maintaining competitive performance \cite{supratak2020tinysleepnet}. DeepSleepNet-Lite further simplifies the architecture and provides calibrated uncertainty estimates \cite{fiorillo2021deepsleepnetlite}. Perslev et al. proposed U-Sleep, a fully convolutional segmentation framework designed for heterogeneous sleep cohorts \cite{perslev2021usleep}. Kuo and Chen investigated hybrid recurrent architectures for large-scale clinical sleep staging \cite{kuo2020automatic}. Temporal convolutional networks have also emerged as an attractive alternative because dilated convolutions provide large receptive fields with relatively few parameters while supporting parallel computation \cite{lea2017temporal, bai2018empirical}. Previous studies have addressed class imbalance using oversampling, cost-sensitive learning, and focal loss \cite{zhou2020automatic, sun2019two, lin2017focal}. Despite these advances, existing lightweight models generally sacrifice classification accuracy to achieve computational efficiency.
Existing studies have significantly advanced automatic sleep stage classification. However, several challenges remain. High-performing models often rely on large convolutional, recurrent, or transformer architectures that are unsuitable for wearable and resource-constrained devices. Lightweight models reduce computational cost but usually compromise classification accuracy. In addition, most methods rely primarily on either learned representations or handcrafted features rather than exploiting their complementary strengths. Finally, the recognition of minority sleep stages, particularly N1, remains challenging because of severe class imbalance. To address these limitations, we propose NanoSleep, a compact hybrid architecture for single-channel EEG sleep stage classification.
\section{Methodology}
\label{sec:method}
\begin{figure*}[!t]
\centerline{\includegraphics[width=0.95\textwidth]{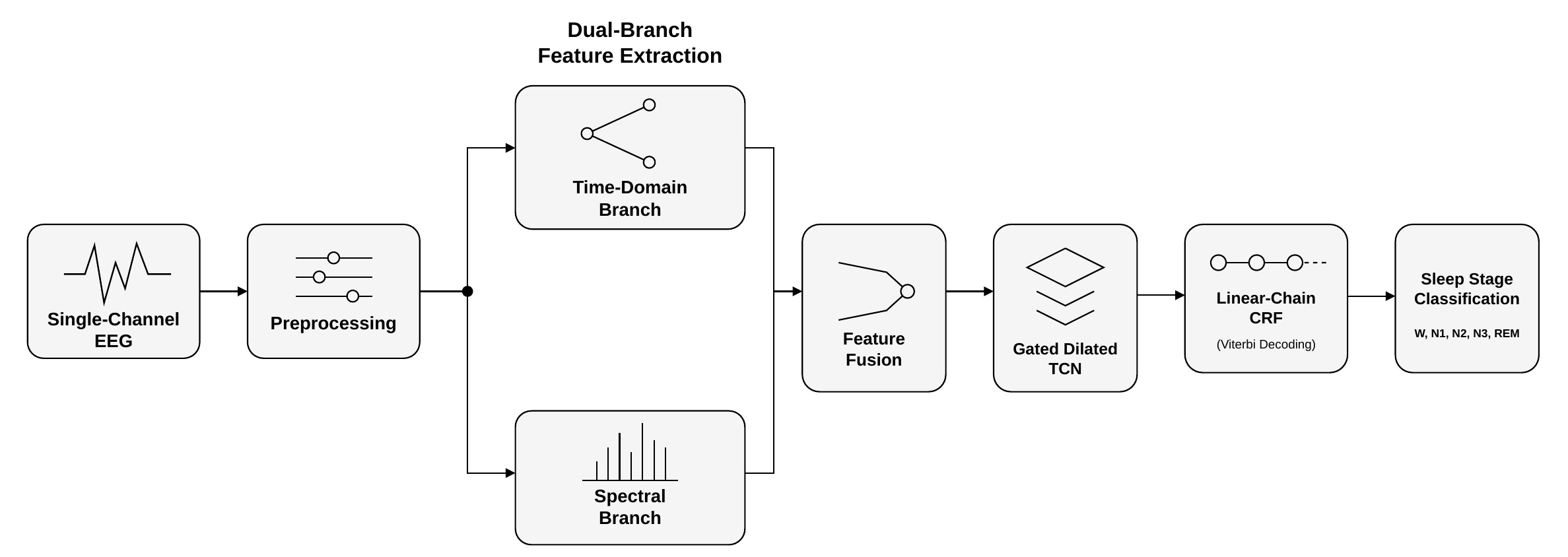}}
\caption{Overall architecture of the proposed NanoSleep framework for single-channel EEG sleep stage classification.}
\label{fig:architecture}
\end{figure*}

In this section, we describe the NanoSleep framework in detail. Fig.~\ref{fig:architecture} shows the complete framework: we preprocess the raw single-channel EEG, extract time-domain and spectral features in parallel, fuse them, process the fusion with a gated dilated TCN backbone, and decode the sleep stage predictions with a CRF. Here, we present the proposed model architecture and the training objective. 
 % The corresponding datasets, cross-validation protocol, hyperparameter values, and evaluation metrics are reported in Section~\ref{sec:setup}. An overview of the complete framework is shown in Fig.~\ref{fig:architecture}.

\subsection{Proposed Model}
\label{subsec:model}
The main contribution of this work is NanoSleep, a compact hybrid model with approximately 0.35 million parameters. We combine learned and interpretable representations, an efficient temporal backbone, and a learnable sequence-level decoder into a single architecture. We design the model to satisfy three key requirements. First, it captures the multi-scale morphology of sleep EEG. Second, it exploits the strong sequential regularities of overnight sleep. Third, it remains compact enough to run on wearable and point-of-care devices. The following subsections describe each component. Each component is now described in turn, and the complete data flow is depicted in Fig.~\ref{fig:architecture}.

\subsubsection{Multi-Scale Time-Domain Branch}
The first branch of the extractor learns time-domain representations directly from the preprocessed signal using a multi-scale convolution module. Sleep EEG contains patterns with different temporal durations. Some events, such as K-complexes, are brief. Others, such as slow oscillations, last several seconds. To capture this diversity, we apply three convolutional kernels in parallel. The small kernel captures short transient events. The medium kernel captures mid-length rhythms, such as sleep spindles. The large kernel captures slow waves. We concatenate the outputs of all three kernels. This operation produces a feature representation that encodes multiple temporal scales and reflects the multi-scale nature of sleep EEG waveforms.

\subsubsection{Spectral Feature Branch}
\begin{figure}[!t]
\centerline{\includegraphics[width=0.48\textwidth]{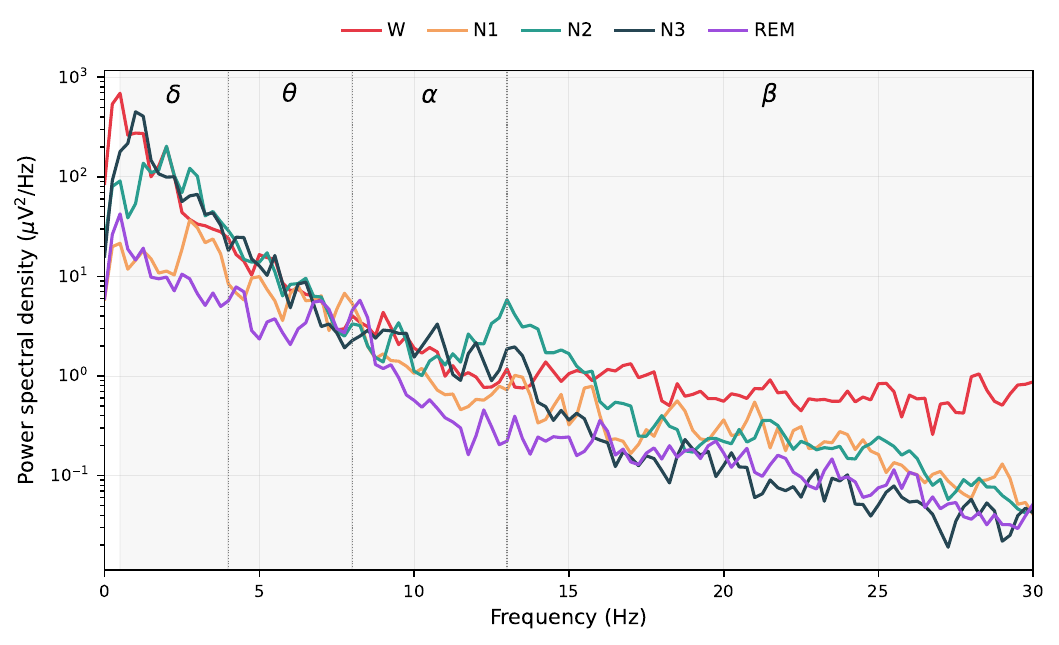}}
\caption{Power spectral density of the single-channel EEG for the five sleep stages, with the standard delta, theta, alpha, and beta frequency bands indicated.}
\label{fig:psd}
\end{figure}

The second branch computes interpretable descriptors that complement the learned representations.
We first compute the relative power spectral density in the standard delta, theta, alpha, and beta frequency bands. Relative power reduces sensitivity to inter-subject amplitude variation.
Fig.~\ref{fig:psd} shows the spectral power distribution for each sleep stage. The figure confirms that these bands contain stage-discriminative information. We also compute multiscale dispersion entropy \cite{rostaghi2016dispersion} and the Hjorth parameters of activity, mobility, and complexity \cite{hjorth1970eeg}. These features describe the signal complexity and temporal dynamics across multiple scales. Next, we pass all descriptors through a small multilayer perceptron, called the spectral multilayer perceptron. This network transforms them into a compact feature vector. We intentionally retain this handcrafted spectral branch alongside the learned branch. The spectral features provide stable and physiologically meaningful information that generalizes well across subjects. The ablation study in Section~\ref{subsec:res_ablation} shows that removing this branch causes one of the largest performance drops.
% It first computes the relative power spectral density of the epoch within the standard delta, theta, alpha, and beta frequency bands, and the use of relative rather than absolute power reduces sensitivity to inter-subject amplitude variation. The distribution of spectral power across these bands is shown for each stage in Fig.~\ref{fig:psd}, which confirms that the bands carry stage-discriminative information. The branch additionally computes multiscale dispersion entropy \cite{rostaghi2016dispersion} and the Hjorth parameters of activity, mobility, and complexity \cite{hjorth1970eeg}, which together quantify the complexity and the dynamic temporal properties of the signal across multiple temporal scales. These descriptors are passed through a small multilayer perceptron, referred to as the spectral multilayer perceptron, that transforms them into a compact feature vector. Retaining a handcrafted spectral branch alongside the learned branch is a deliberate design choice. The spectral descriptors provide a stable, physiologically grounded floor for the model that generalizes well across subjects, and the ablation study in Section~\ref{subsec:res_ablation} confirms that removing this branch causes one of the largest performance drops in the model.

\subsubsection{Feature Fusion}
The fusion stage combines the learned time-domain features with the transformed spectral and complexity features. We concatenate the outputs of both branches to form a unified feature representation. This design allows the model to leverage the flexibility of learned features and the robustness of handcrafted descriptors. It also preserves the physiological interpretability of the spectral features while improving the overall representation.

\subsubsection{Gated Dilated Temporal Convolutional Backbone}
The fused features are processed by a gated dilated temporal convolutional network, which serves as the sequence backbone of NanoSleep. We use dilated convolutions to capture long-range temporal context across many epochs. This design increases the receptive field without adding many parameters or requiring sequential computation \cite{yu2016multiscale, bai2018empirical}. Each layer includes a gated linear unit \cite{dauphin2017language}. The gate learns which temporal information to retain and which to suppress. We then apply a squeeze-and-excitation block \cite{hu2018squeeze} to recalibrate the feature channels. This block emphasizes informative channels and reduces the influence of less useful ones. Together, dilation, gating, and channel recalibration produce a compact and expressive backbone.
% The fused representation is processed by a gated dilated temporal convolutional network, which serves as the sequence backbone of NanoSleep. Dilated convolutions enlarge the receptive field exponentially with depth, so that long-range temporal context across many epochs is captured without a large parameter count and without the sequential computation that is required by recurrent networks \cite{yu2016multiscale, bai2018empirical}. Each layer of the backbone is equipped with a gated linear unit \cite{dauphin2017language}, which acts as a learned filter that decides which temporal information is propagated and which is suppressed. A squeeze-and-excitation block \cite{hu2018squeeze} is then applied to recalibrate the feature channels, so that the channels that are most informative for the current epoch are emphasized and the less informative channels are attenuated. The combination of dilation, gating, and channel recalibration yields an expressive yet compact backbone, and it is the principal reason for the favourable accuracy-to-size trade-off reported in Section~\ref{sec:results}.

\subsubsection{Sequence-Level Decoding}
Sleep stages follow well-established physiological transition patterns. We refine the backbone predictions with a learnable linear-chain conditional random field (CRF) \cite{lafferty2001conditional}. The CRF learns a $5\times5$ transition matrix that captures common stage transitions. For example, a direct transition from deep sleep to REM sleep is uncommon. For an input sequence and a candidate label sequence $y$, the CRF assigns the following score:
\begin{equation}
s(y) = \sum_{t=1}^{T} U_{t}(y_{t}) + \sum_{t=1}^{T-1} A(y_{t}, y_{t+1}),
\label{eq:crf}
\end{equation}
where $U_{t}(y_{t})$ is the unary score that the backbone assigns to stage $y_{t}$ at epoch $t$, $A(y_{t}, y_{t+1})$ is the learned transition score between consecutive stages, and $T$ is the number of epochs in the sequence. During inference, we use the Viterbi algorithm \cite{viterbi1967error} to find the label sequence with the highest score. This decoding step removes implausible isolated predictions. For example, it suppresses a single wake epoch that appears within a long period of deep sleep. As a result, the final predictions become more consistent with the natural progression of sleep stages

\subsection{Training Objective}
\label{subsec:training}
We design the training objective to address class imbalance and improve predictions at sleep stage transitions. The objective combines a weighted calibrated focal loss with the negative log-likelihood of the conditional random field (CRF).
The focal loss \cite{lin2017focal} reduces the contribution of easy examples and focuses training on difficult epochs, such as those belonging to the N1 stage.
 For an epoch with ground-truth stage $c$ and predicted probability $p_{c}$, the per-epoch focal loss is
\begin{equation}
\mathcal{L}_{\mathrm{focal}} = - w_{c} \, (1 - p_{c})^{\gamma_{c}} \, \log(p_{c}),
\label{eq:focal}
\end{equation}
where $w_{c}$ is the class weight and $\gamma_{c}$ is the focusing parameter of class $c$. We compute the class weights using a sub-linear function of the inverse class frequency. This strategy increases the importance of rare stages without neglecting the majority classes. We also use a class-specific focusing schedule. We assign larger focusing parameters to the difficult N1 and N3 stages and smaller values to the majority stages.

% The class weights in \eqref{eq:focal} are obtained from a sub-linear function of the inverse class frequency, which up-weights rare stages without allowing the dominant stages to be ignored. A per-class focusing schedule is adopted, in which a larger focusing parameter is assigned to the difficult N1 and N3 stages and a smaller value is assigned to the majority stages. To discourage overconfident misclassifications, the predicted probabilities are calibrated through probability smoothing, which encourages the output of the model to reflect its true confidence \cite{guo2017calibration}. A boundary-aware penalty additionally multiplies the loss in \eqref{eq:focal} by a factor of two whenever a stage transition occurs between consecutive epochs, which forces the model to be precise at the borders of sleep cycles. The overall objective is the sum of the weighted calibrated focal loss and the negative log-likelihood of the conditional random field, the latter being computed from the score in 

We calibrate the predicted probabilities through probability smoothing. This step discourages overconfident predictions and produces confidence scores that better reflect the model's true uncertainty \cite{guo2017calibration}. We further apply a boundary-aware penalty. We double the focal loss whenever a stage transition occurs between consecutive epochs. This penalty encourages the model to make more accurate predictions at sleep stage boundaries. The final training objective is the sum of the weighted calibrated focal loss and the CRF negative log-likelihood. We compute the CRF loss from the sequence score in
\eqref{eq:crf}.

\section{Experimental Setup}
\label{sec:setup}

In this section, we describe the experimental setup, including the dataset, preprocessing steps, data augmentation strategy, experimental configuration, and evaluation metrics used for sleep stage classification.
% This section describes the experimental setup used to evaluate the NanoSleep framework. It reports the two public datasets and the reasons for their selection, the subject-wise cross-validation protocol, the complete set of implementation and training details for reproducibility, and the evaluation metrics with their corresponding equations.

\subsection{Datasets and Training Protocol}
\label{subsec:setup_data}
\begin{figure}[!t]
\centerline{\includegraphics[width=0.48\textwidth]{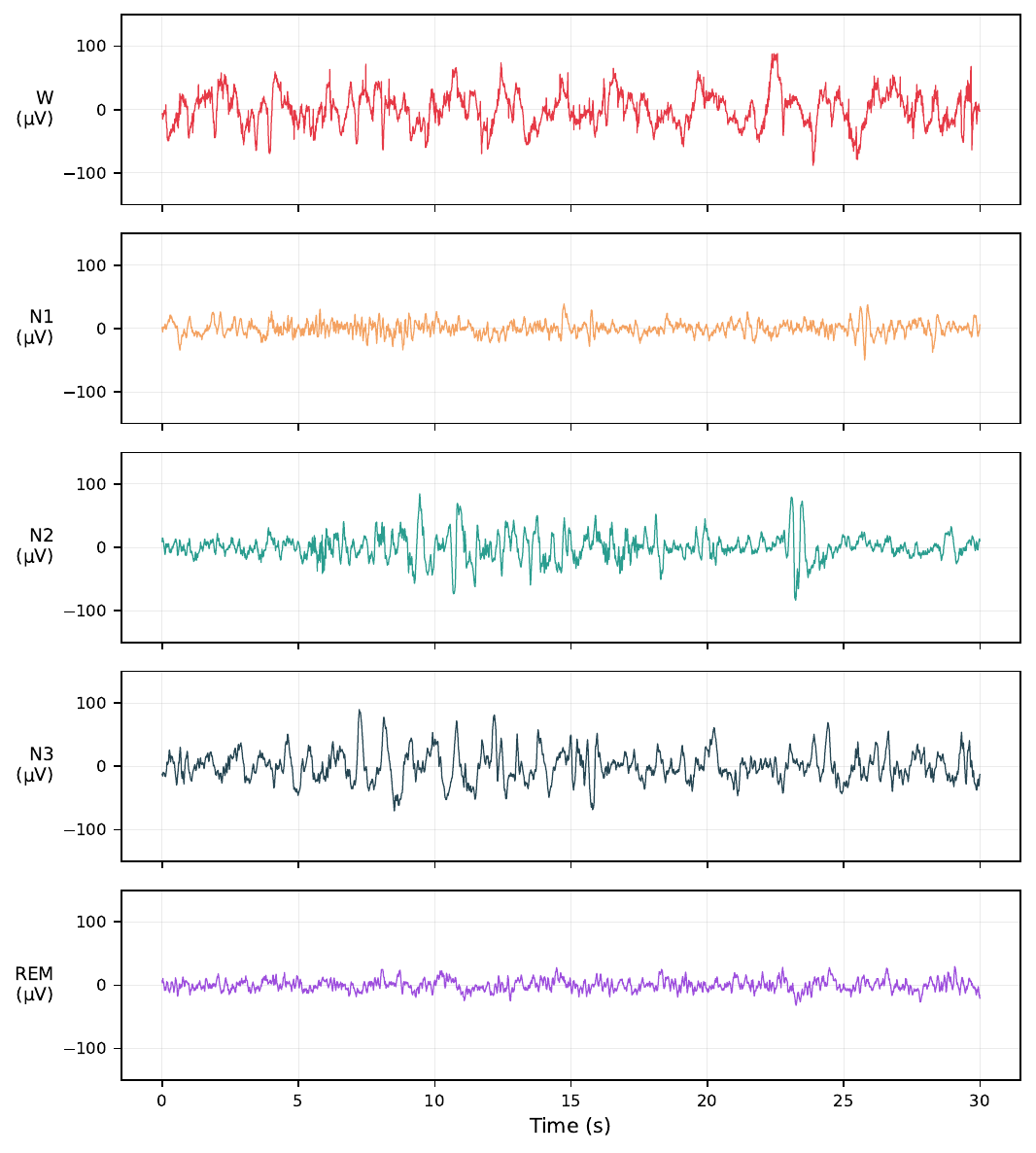}}
\caption{Representative 30\,s single-channel EEG epochs from the Fpz-Cz channel of the Sleep-EDF dataset for the five sleep stages. The N2 epoch exhibits sleep spindles and K-complexes, the N3 epoch is dominated by high-amplitude slow waves, and the REM epoch presents a low-amplitude mixed-frequency pattern.}
\label{fig:rawepochs}
\end{figure}

\begin{figure}[!t]
\centerline{\includegraphics[width=0.48\textwidth]{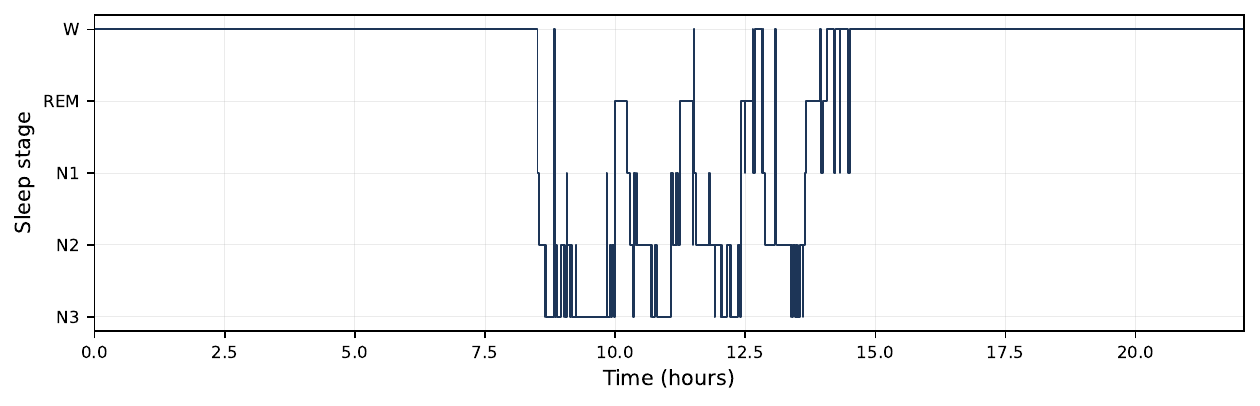}}
\caption{Full-night hypnogram of a representative subject from the Sleep-EDF dataset, illustrating the cyclic alternation of sleep stages across the night and the comparatively small proportion of the N1 stage.}
\label{fig:hypnogram}
\end{figure}

We evaluate the proposed model and benchmark methods on two publicly available sleep EEG datasets. These datasets include recordings from both healthy individuals and subjects with sleep disorders. This diversity allows us to assess both classification accuracy and generalization. The Sleep-EDF dataset and the Sleep-EDF-Expanded dataset are available through the PhysioNet repository \cite{sleepedfexpanded2018, pollard2026physionet, kemp2000analysis}.
The first dataset is Sleep-EDF. We use the Fpz-Cz EEG channel from overnight recordings sampled at 100~Hz. The second dataset is Sleep-EDF-Expanded. We again use the Fpz-Cz channel. This dataset contains many more subjects and recordings, making it a stronger benchmark for evaluating robustness and cross-subject generalization.
% The first dataset is the classical Sleep-EDF dataset, from which the Fpz-Cz EEG channel was used. It is a long-standing benchmark that consists of overnight recordings sampled at 100~Hz. The second dataset is the larger Sleep-EDF-Expanded dataset, again using the Fpz-Cz channel, which contains a substantially greater number of subjects and recordings and therefore provides a stronger test of robustness and of generalization across individuals.
We segment all recordings into 30,s epochs according to the AASM guidelines. We assign each epoch to one of five sleep stages: W, N1, N2, N3, or REM. For datasets that use the older scoring standard, we merge the N3 and N4 stages into a single N3 class.
Table~\ref{tab:datadist} summarizes the number of subjects, EEG channels, and sampling rates for all datasets. Fig.~\ref{fig:rawepochs} shows representative EEG epochs for each sleep stage, and Fig.~\ref{fig:hypnogram} shows a complete overnight hypnogram, both drawn from the Sleep-EDF dataset.

% All recordings were segmented into 30\,s epochs in accordance with the AASM guidelines, and each epoch was mapped to one of five stages, namely W, N1, N2, N3, and REM, with the older N3 and N4 labels merged into a single N3 stage. The number of subjects, the EEG channel, and the sampling rate of each dataset are summarized in Table~\ref{tab:datadist}. Fig.~\ref{fig:rawepochs} shows representative 30\,s epochs of each stage, and Fig.~\ref{fig:hypnogram} shows a complete overnight hypnogram, both drawn from the Sleep-EDF dataset.

\begin{table}[!t]
\caption{Composition of the two datasets used in this study.}
\label{tab:datadist}
\centering
\begin{tabular}{lcc}
\hline
\textbf{Property} & \textbf{Sleep-EDF} & \textbf{Sleep-EDF-Exp.} \\
\hline
Subjects & 20 & 78 \\
EEG channel & Fpz-Cz & Fpz-Cz \\
Sampling rate & 100~Hz & 100~Hz \\
\hline
\end{tabular}
\end{table}

% To obtain an unbiased estimate of generalization to unseen individuals, all experiments used subject-wise cross-validation, in which the recordings of any given subject appear exclusively in either the training set or the test set within a fold. A 20-fold CV scheme was used for Sleep-EDF and a 10-fold CV scheme for Sleep-EDF-Expanded, with the number of folds chosen according to the size of each dataset. Within each training fold, a further subset of subjects was held out as a validation split for model selection and for the tuning of hyperparameters.
To evaluate generalization, we perform subject-wise cross-validation. We ensure that recordings from the same subject never appear in both the training and test sets within a fold. We use 20-fold cross-validation for Sleep-EDF and 10-fold cross-validation for Sleep-EDF-Expanded. We select the number of folds according to the size of each dataset. Within every training fold, we reserve a subset of subjects for validation. We use this validation set for model selection and hyperparameter tuning.

\subsection{Preprocessing Pipeline}
\label{subsec:preproc}
% The preprocessing pipeline reduces noise and exposes stage-relevant frequency content while keeping transient waveforms experts require. The three phases are learnable spectral filtering, robust normalization, and wavelet denoising. It comprises three stages, namely learnable spectral filtering, robust normalization, and wavelet denoising.
We preprocess the EEG signals to reduce noise while preserving stage-specific patterns. The preprocessing pipeline consists of three stages: learnable spectral filtering, robust normalization, and wavelet denoising.

% The first stage replaces the fixed band-pass filter that is common in earlier pipelines with a learnable Sinc-convolutional layer \cite{ravanelli2018speaker}. In a conventional pipeline the cutoff frequencies of the band-pass filter are chosen heuristically and remain fixed, which may discard frequency content that is informative for a particular cohort or recording montage. The Sinc-convolutional layer instead parameterizes each filter by a low and a high cutoff frequency and learns these frequencies jointly with the rest of the network. 
First, we replace the fixed band-pass filter used in conventional pipelines with a learnable Sinc-convolutional layer \cite{ravanelli2018speaker}. Traditional band-pass filters use manually selected cutoff frequencies that remain fixed during training. As a result, they may remove useful frequency information. In contrast, the Sinc-convolutional layer learns the cutoff frequencies directly from the data. Each filter is defined by a low cutoff frequency $f_{1}$ and a high cutoff frequency $f_{2}$. The filter response is constructed from the difference of two sinc functions and
% The filter is realized as a band-pass response constructed from the difference of two sinc functions, and
a filter $g$ is defined in the time domain as
\begin{equation}
g[n, f_{1}, f_{2}] = 2 f_{2}\,\mathrm{sinc}(2\pi f_{2} n) - 2 f_{1}\,\mathrm{sinc}(2\pi f_{1} n),
\label{eq:sinc}
\end{equation}
% where $f_{1}$ and $f_{2}$ are the learnable low and high cutoff frequencies, $n$ is the sample index, and $\mathrm{sinc}(x) = \sin(x)/x$. 
where $n$ is the sample index and $\mathrm{sinc}(x)=\sin(x)/x$.
We learn only the two cutoff frequencies for each filter. This design keeps the number of trainable parameters low while allowing the network to discover the most informative frequency bands for sleep staging.

% Because only the two cutoff frequencies of each filter are learned, the layer in \eqref{eq:sinc} introduces very few parameters while allowing the network to discover the frequency bands that are most informative for sleep staging.

% The second stage normalizes each epoch using the median and the median absolute deviation rather than the mean and the standard deviation. The use of robust statistics is motivated by the fact that EEG epochs frequently contain transient high-amplitude artefacts, and that the mean and the standard deviation are sensitive to such outliers. Given an epoch $x$, the normalized epoch $\tilde{x}$ is computed as
% \begin{equation}
% \tilde{x} = \frac{x - \mathrm{median}(x)}{\mathrm{MAD}(x) + \epsilon},
% \label{eq:madnorm}
% \end{equation}
% where $\mathrm{MAD}(x)$ denotes the median absolute deviation and $\epsilon$ is a small constant that prevents division by zero. The normalization in \eqref{eq:madnorm} reduces the influence of amplitude outliers and of inter-subject amplitude variability, and it thereby improves the comparability of epochs that originate from different subjects and recording sessions.
Next, we normalize each EEG epoch using the median and the median absolute deviation (MAD) instead of the mean and standard deviation. EEG recordings often contain high-amplitude artifacts that distort conventional normalization. Robust statistics reduce the influence of these outliers. For an epoch $x$, we compute the normalized signal as
\begin{equation}
\tilde{x} = \frac{x - \mathrm{median}(x)}{\mathrm{MAD}(x) + \epsilon},
\label{eq}
\end{equation}
where $\mathrm{MAD}(x)$ is the median absolute deviation and $\epsilon$ is a small constant that prevents division by zero. This normalization reduces inter-subject amplitude variation and improves the consistency of the input signals.
Finally, we apply wavelet denoising using the Daubechies-4 wavelet with soft thresholding \cite{donoho1994ideal}. We first decompose the EEG signal into wavelet coefficients. We then apply soft thresholding to the detail coefficients and reconstruct the signal. This process removes low-amplitude noise while preserving clinically important waveforms, such as sleep spindles and K-complexes, which are essential for identifying the N2 stage.
% The third stage applies wavelet denoising using the Daubechies-4 wavelet with soft thresholding \cite{donoho1994ideal}. The EEG signal is decomposed into wavelet coefficients, a soft threshold is applied to the detail coefficients, and the signal is reconstructed. Soft thresholding suppresses low-amplitude transient noise while preserving the sleep spindles and K-complexes that are diagnostic of the N2 stage, because these waveforms are represented compactly by a small number of large wavelet coefficients and are therefore retained by the threshold.

\subsection{Data Augmentation}
\label{subsec:augment}

\begin{figure}[!t]
\centerline{\includegraphics[width=0.48\textwidth]{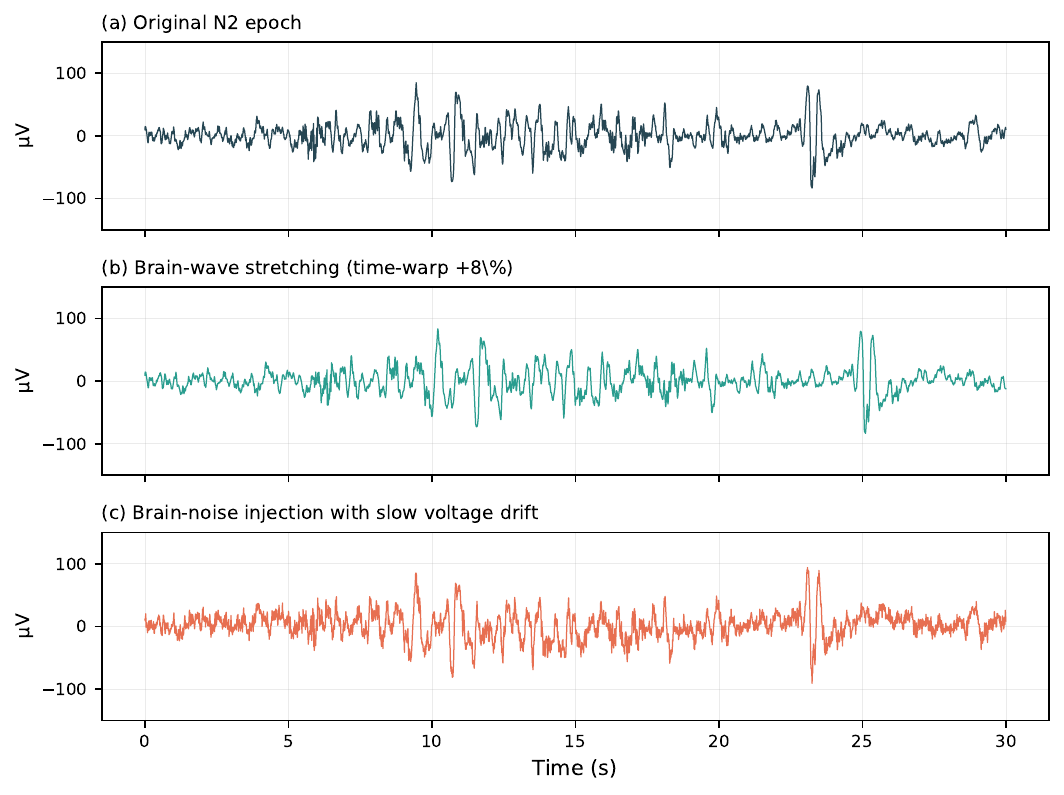}}
\caption{Effect of the data augmentation operations on a representative N2 epoch. Panel (a) shows the original epoch, panel (b) shows the epoch after brain-wave stretching, and panel (c) shows the epoch after brain-noise injection with a slow voltage drift.}
\label{fig:augmentation}
\end{figure}

We apply three data augmentation techniques during training to improve generalization and reduce the effect of class imbalance. These augmentations increase the diversity of the training data, especially for minority sleep stages. First, we apply brain-wave stretching. We slightly stretch or compress the time axis of each EEG epoch. This operation simulates the natural variation of brain rhythms across subjects and sleep cycles. Second, we apply brain-noise injection. We add realistic broadband noise and slow baseline drifts to the EEG signal. This augmentation mimics background brain activity and recording artifacts commonly observed in real-world sleep studies. Third, we use a contextual augmentation based on feature-space CutMix \cite{yun2019cutmix}. We exchange segments of the latent representations between sequences that belong to the same sleep stage. This operation encourages the model to learn more robust stage representations and improves discrimination at stage boundaries. Fig.~\ref{fig:augmentation} illustrates the effects of brain-wave stretching and brain-noise injection on a representative N2 epoch.
% To improve generalization and to counteract the limited number of minority-stage epochs, three complementary data augmentation operations are applied during training. The first operation, referred to as brain-wave stretching, slightly stretches or compresses the time axis of an epoch in order to mimic the natural variability of human EEG rhythms across subjects and across sleep cycles. The second operation, referred to as brain-noise injection, adds realistic broadband noise together with slow voltage drifts, so as to reproduce the background activity and the baseline wander that are characteristic of a living brain and of practical recording conditions. The third operation is a contextual augmentation based on a feature-space variant of CutMix \cite{yun2019cutmix}, in which segments of the latent representation are exchanged between sequences of the same class in order to sharpen the boundaries between stages. The effect of the first two operations on a representative N2 epoch is illustrated in Fig.~\ref{fig:augmentation}.

\subsection{Experimental Configuration and Reproducibility}
\label{subsec:setup_impl}
This subsection describes the implementation and training settings used in our experiments. These details allow the proposed method to be reproduced.

We use EEG signals sampled at 100~Hz. Each 30,s epoch therefore contains 3000 samples. The Sinc-convolutional front end contains 32 learnable band-pass filters, each with a length of 65 samples. We initialize the filters to cover the 0.5--40~Hz frequency range, which includes the main sleep rhythms. We perform wavelet denoising using the Daubechies-4 wavelet with four decomposition levels and a universal soft threshold.

The time-domain branch uses three parallel one-dimensional convolutional layers with kernel sizes of 7, 25, and 51 samples. Each branch produces 32 feature maps to capture short-, medium-, and long-duration temporal patterns. The spectral multilayer perceptron contains two hidden layers with 64 and 32 neurons, respectively. Both layers use rectified linear unit (ReLU) activations. The temporal backbone consists of six residual blocks with dilation rates of 1, 2, 4, 8, 16, and 32. Each block uses a kernel size of 7 and 64 feature channels. Each residual block includes a gated linear unit, batch normalization \cite{ioffe2015batch}, and dropout with a rate of 0.3 \cite{srivastava2014dropout}. The squeeze-and-excitation module uses a channel reduction ratio of 8.

We train the model using the Adam optimizer \cite{kingma2015adam}. The initial learning rate is $1\times10^{-3}$ with cosine-annealing scheduling and a weight decay of $1\times10^{-4}$. We use a batch size of 32 sequences, where each sequence contains 20 consecutive epochs. We train the model for up to 120 epochs and apply early stopping if the validation performance does not improve for 15 consecutive epochs. For the focal loss in \eqref{eq:focal}, we set the focusing parameter to 2.5 for the N1 and N3 stages and 1.0 for the W, N2, and REM stages. We set the probability-smoothing factor to 0.1. We apply each data augmentation method described in Section~\ref{subsec:augment} online with a probability of 0.5.

The complete model contains approximately 0.35 million trainable parameters. We implement all experiments in PyTorch and run them on a single NVIDIA RTX~3090 GPU with 24~GB of memory. We fix the random seed to 42 for every cross-validation fold to ensure reproducible data splits, model initialization, and training.

\subsection{Evaluation Metrics}
\label{subsec:setup_metrics}
We evaluate NanoSleep using several performance metrics that measure overall accuracy, per-stage performance, and agreement with expert annotations. These metrics are widely used in automatic sleep staging studies \cite{supratak2017deepsleepnet, zhou2020automatic}. Let $\mathrm{TP}$, $\mathrm{TN}$, $\mathrm{FP}$, and $\mathrm{FN}$ denote the numbers of true positive, true negative, false positive, and false negative epochs, respectively.

We first report the overall accuracy (ACC), which measures the proportion of correctly classified epochs:
\begin{equation}
\mathrm{ACC} = \frac{\mathrm{TP} + \mathrm{TN}}{\mathrm{TP} + \mathrm{TN} + \mathrm{FP} + \mathrm{FN}}.
\label{eq}
\end{equation}
Accuracy provides an overall measure of performance. However, it can be biased toward majority sleep stages in imbalanced datasets.

We therefore report precision and recall for each sleep stage. Precision measures how many predicted epochs are correct, while recall measures how many true epochs are correctly identified:
\begin{equation}
\mathrm{Precision} = \frac{\mathrm{TP}}{\mathrm{TP} + \mathrm{FP}}, \qquad
\mathrm{Recall} = \frac{\mathrm{TP}}{\mathrm{TP} + \mathrm{FN}}.
\label{eq}
\end{equation}

We also compute the F1-score, which balances precision and recall:
\begin{equation}
\mathrm{F1} = \frac{2 \times \mathrm{Precision} \times \mathrm{Recall}}{\mathrm{Precision} + \mathrm{Recall}}.
\label{eq}
\end{equation}

To account for class imbalance, we report the macro F1-score (MF1). This metric computes the average F1-score across all sleep stages:
\begin{equation}
\mathrm{MF1} = \frac{1}{C} \sum_{i=1}^{C} \mathrm{F1}{i},
\label{eq}
\end{equation}
where $C=5$ is the number of sleep stages and $\mathrm{F1}{i}$ is the F1-score for stage $i$. Since each stage contributes equally, MF1 reflects the model's performance on both majority and minority classes.

Finally, we report Cohen's kappa coefficient ($\kappa$), which measures the agreement between the model predictions and expert annotations after correcting for chance agreement \cite{cohen1960coefficient}:
\begin{equation}
\kappa = \frac{p_{o} - p_{e}}{1 - p_{e}},
\label{eq}
\end{equation}
where $p_{o}$ is the observed agreement and $p_{e}$ is the expected agreement by chance. A $\kappa$ value above 0.80 indicates outstanding agreement, while values between 0.61 and 0.80 indicate substantial agreement. We also present confusion matrices to visualize the classification performance and error patterns for each sleep stage on every dataset.

\section{Results and Discussion}
\label{sec:results}
In this section, we evaluate the performance of NanoSleep on two benchmark sleep EEG datasets. We compare the proposed model with representative state-of-the-art methods under the same subject-wise cross-validation protocol. We then analyze the confusion matrices to examine stage-level errors and perform an ablation study to quantify the contribution of each component. Finally, we discuss the main findings and their implications for compact and accurate sleep stage classification.

\subsection{Performance on the Sleep-EDF Dataset}
\label{subsec:res_sleepedf}
We first evaluate NanoSleep on the Sleep-EDF dataset. This experiment assesses whether the proposed architecture improves classification performance while maintaining a compact model size. We compare NanoSleep with six representative baseline models using 20-fold subject-wise cross-validation. We report overall accuracy (ACC), macro F1-score (MF1), Cohen's kappa ($\kappa$), per-stage F1-score, and model size. Table~\ref{tab:sleepedf} summarizes the results. NanoSleep achieves the best overall performance, with an accuracy of 86.5\%, a macro F1-score of 82.1\%, and a kappa coefficient of 0.81. It outperforms TinySleepNet by 1.1\% in accuracy, 1.6\% in macro F1-score, and 0.01 in kappa. NanoSleep also uses only 0.35~M parameters, compared with 1.3~M for TinySleepNet. Thus, it requires only 27\% of the parameters while achieving higher performance.

NanoSleep also outperforms larger and more complex models. It improves accuracy by 2.2\% over SleepEEGNet, 4.6\% over DeepSleepNet, 2.6\% over XSleepNet, and 2.1\% over AttnSleep. It also improves the macro F1-score by 2.4\%, 5.5\%, 3.4\%, and 4.0\%, respectively. These results show that NanoSleep consistently outperforms recurrent, attention-based, lightweight, and high-capacity architectures. At the stage level, NanoSleep achieves the highest F1-scores for W, N2, and REM, with values of 94.4\%, 89.8\%, and 87.0\%, respectively. It improves W recognition by 4.3\% over TinySleepNet and 5.2\% over SleepEEGNet. It also achieves the best performance on N2, the dominant sleep stage, with a 1.3\% improvement over TinySleepNet. For REM, NanoSleep exceeds the strongest competing model by 1.6\%. AttnSleep achieves the highest F1-score for N3, but NanoSleep remains within 1.6\% while providing better overall accuracy, macro F1-score, kappa, and a substantially smaller model. N1 is the only stage where NanoSleep is not the best. Its F1-score of 50.8\% remains close to the highest reported value of 52.9\%.

We next analyze the confusion matrix to identify the remaining classification errors. Figure~\ref{fig:confusion_matrices}(a) shows that W, N2, N3, and REM are classified with high reliability. Their recall values reach 94.0\%, 90.5\%, 89.0\%, and 88.0\%, respectively. In contrast, N1 remains the most challenging stage, with a recall of 47.5\%. Most N1 errors occur in the neighboring stages. Specifically, 26.5\% of N1 epochs are classified as N2, 14.0\% as REM, and 11.5\% as W. This behavior agrees with sleep physiology because N1 is a short transitional stage that shares characteristics with adjacent stages and often produces disagreement among expert scorers \cite{rosenberg2013american}. We also observe that 10.7\% of N3 epochs are classified as N2. This confusion reflects the gradual transition between these two stages.

Overall, these results demonstrate that NanoSleep achieves an excellent accuracy-to-parameter trade-off. The hybrid feature extractor combines complementary temporal and spectral information, while the temporal backbone captures long-range sleep dependencies efficiently. Together, these components enable NanoSleep to outperform both lightweight and high-capacity models while maintaining a compact architecture.

\begin{table*}[!t]
\caption{Performance comparison on the Sleep-EDF dataset under 20-fold subject-wise cross-validation.}
\label{tab:sleepedf}
\centering
\resizebox{\textwidth}{!}{%
\begin{tabular}{lccccccccc}
\hline
\textbf{Model} & \textbf{ACC (\%)} & \textbf{MF1 (\%)} & \textbf{$\kappa$} & \textbf{W} & \textbf{N1} & \textbf{N2} & \textbf{N3} & \textbf{REM} & \textbf{Param.} \\
\hline
SleepEEGNet & 84.3 $\pm$ 0.4 & 79.7 $\pm$ 0.8 & 0.79 $\pm$ 0.02 & 89.2 $\pm$ 0.5 & 52.2 $\pm$ 0.8 & 86.8 $\pm$ 0.3 & 85.1 $\pm$ 0.3 & 85.0 $\pm$ 0.3 & 2.6\,M \\
DeepSleepNet & 81.9 $\pm$ 0.3 & 76.6 $\pm$ 0.3 & 0.76 $\pm$ 0.04 & 86.7 $\pm$ 0.4 & 45.5 $\pm$ 0.3 & 85.1 $\pm$ 0.3 & 83.3 $\pm$ 0.4 & 82.6 $\pm$ 0.2 & 24.7\,M \\
XSleepNet & 83.9 $\pm$ 0.3 & 78.7 $\pm$ 0.2 & 0.77 $\pm$ 0.04 & 81.6 $\pm$ 0.4 & \textbf{52.9 $\pm$ 0.3} & 88.1 $\pm$ 0.2 & 85.3 $\pm$ 0.3 & 85.4 $\pm$ 0.3 & 5.8\,M \\
AttnSleep & 84.4 $\pm$ 0.3 & 78.1 $\pm$ 0.3 & 0.79 $\pm$ 0.03 & 89.7 $\pm$ 0.3 & 42.6 $\pm$ 0.3 & 88.8 $\pm$ 0.3 & \textbf{90.2 $\pm$ 0.3} & 79.0 $\pm$ 0.4 & 0.52\,M \\
DeepSleepNet-Lite & 84.0 $\pm$ 0.4 & 78.0 $\pm$ 0.3 & 0.78 $\pm$ 0.04 & 87.1 $\pm$ 0.4 & 44.4 $\pm$ 0.3 & 87.9 $\pm$ 0.2 & 88.2 $\pm$ 0.4 & 82.4 $\pm$ 0.3 & 0.6\,M \\
TinySleepNet & 85.4 $\pm$ 0.3 & 80.5 $\pm$ 0.3 & 0.80 $\pm$ 0.04 & 90.1 $\pm$ 0.3 & 51.4 $\pm$ 0.3 & 88.5 $\pm$ 0.2 & 88.3 $\pm$ 0.3 & 84.3 $\pm$ 0.4 & 1.3\,M \\
\textbf{NanoSleep} & \textbf{86.5 $\pm$ 0.3} & \textbf{82.1 $\pm$ 0.3} & \textbf{0.81 $\pm$ 0.04} & \textbf{94.4 $\pm$ 0.4} & 50.8 $\pm$ 0.3 & \textbf{89.8 $\pm$ 0.4} & 88.6 $\pm$ 0.4 & \textbf{87.0 $\pm$ 0.3} & \textbf{0.35\,M} \\
\hline
\end{tabular}
}
\vspace{2pt}
\parbox{\textwidth}{\footnotesize \textit{Note:} Results are reported as mean $\pm$ standard deviation over 20 folds. Bold indicates the best result in each column. Statistical significance was evaluated using a two-sided paired Wilcoxon signed-rank test, with a $p$-value $<0.05$ considered statistically significant.}
\end{table*}

\subsection{Performance on the Sleep-EDF-Expanded Dataset}
\label{subsec:res_sleepedfx}
We next evaluate NanoSleep on the larger Sleep-EDF-Expanded dataset to assess its robustness on a more heterogeneous population. We perform 10-fold subject-wise cross-validation and compare NanoSleep with the same baseline models.Table~\ref{tab:sleepedfx} summarizes the results. NanoSleep again achieves the best overall performance, with an accuracy of 84.3\%, a macro F1-score of 79.8\%, and a kappa coefficient of 0.78. It improves upon TinySleepNet by 1.2\% in accuracy, 1.6\% in macro F1-score, and 0.01 in kappa while using only 0.35~M parameters, compared with 1.3~M. These results show that the performance gains observed on the original Sleep-EDF dataset remain consistent on a larger and more diverse cohort.

NanoSleep also outperforms the remaining baseline models. It improves accuracy by 4.3\% over SleepEEGNet, 6.5\% over DeepSleepNet, 4.0\% over XSleepNet, 3.0\% over AttnSleep, and 4.0\% over DeepSleepNet-Lite. It also improves the macro F1-score by 6.2\%, 8.0\%, 3.4\%, 4.6\%, and 4.6\%, respectively. These results demonstrate that NanoSleep generalizes well across different model families while maintaining a substantially smaller parameter count. At the stage level, NanoSleep achieves the highest F1-scores for W, N1, N2, and REM, with values of 94.7\%, 51.8\%, 87.0\%, and 84.2\%, respectively. It consistently improves W recognition across both Sleep-EDF datasets, indicating effective separation between wakefulness and sleep. NanoSleep also achieves the best N1 performance, although this stage remains the most challenging. The weighted focal loss and contextual feature learning contribute to this improvement. For N2, NanoSleep improves the F1-score by 1.7\% over TinySleepNet and 2.0\% over AttnSleep. For REM, it improves by 3.9\% over TinySleepNet and 10.0\% over AttnSleep. AttnSleep achieves the highest F1-score for N3, but NanoSleep remains within 0.7\% while achieving higher overall accuracy, macro F1-score, kappa, and better performance on the remaining stages. These consistent improvements indicate that NanoSleep generalizes well across subjects and does not overfit the smaller Sleep-EDF dataset.

\begin{figure*}[!t]
\centering
\begin{subfigure}[t]{0.48\textwidth}
\centering
\includegraphics[width=\linewidth]{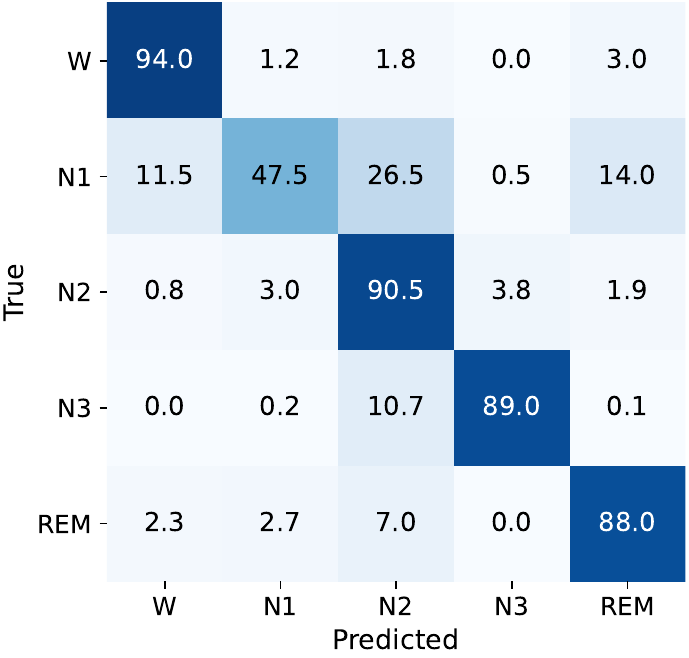}
\caption{Sleep-EDF dataset.}
\label{fig:cm_sleepedf}
\end{subfigure}
\hfill
\begin{subfigure}[t]{0.48\textwidth}
\centering
\includegraphics[width=\linewidth]{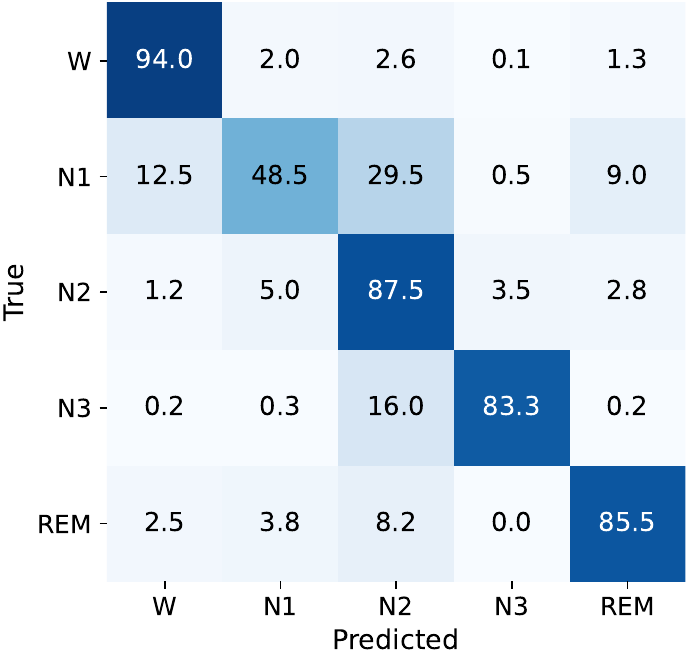}
\caption{Sleep-EDF-Expanded dataset.}
\label{fig:cm_sleepedfx}
\end{subfigure}
\caption{Confusion matrices of NanoSleep on the two evaluation datasets. Rows denote the expert annotation and columns denote the model prediction.}
\label{fig:confusion_matrices}
\end{figure*}

We next analyze the confusion matrix to examine the remaining classification errors. Figure~\ref{fig:confusion_matrices}(b) shows that W, N2, and REM are recognized with high reliability. In contrast, N1 remains the most difficult stage, with a recall of 48.5\%. Most N1 errors occur in neighboring stages, which is expected because N1 represents a transitional sleep stage. We also observe greater confusion between N3 and N2 than on the original Sleep-EDF dataset. Specifically, 16.0\% of N3 epochs are classified as N2. This pattern is consistent with the broader age range and greater inter-subject variability of the Sleep-EDF-Expanded cohort, where reduced slow-wave activity makes the boundary between N2 and N3 less distinct.

Overall, these results demonstrate that NanoSleep maintains high accuracy and strong generalization on a larger and more heterogeneous dataset. The model consistently outperforms both lightweight and high-capacity baselines while using substantially fewer parameters. These findings support NanoSleep as an efficient and robust solution for automatic sleep stage classification.

\begin{table*}[!t]
\caption{Performance comparison on the Sleep-EDF-Expanded dataset under 10-fold subject-wise cross-validation.}
\label{tab:sleepedfx}
\centering
\resizebox{\textwidth}{!}{%
\begin{tabular}{lccccccccc}
\hline
\textbf{Model} & \textbf{ACC (\%)} & \textbf{MF1 (\%)} & \textbf{$\kappa$} & \textbf{W} & \textbf{N1} & \textbf{N2} & \textbf{N3} & \textbf{REM} & \textbf{Param.} \\
\hline
SleepEEGNet & 80.0 $\pm$ 0.5 & 73.6 $\pm$ 0.5 & 0.73 $\pm$ 0.03 & 91.7 $\pm$ 0.5 & 44.1 $\pm$ 1.7 & 82.5 $\pm$ 0.6 & 73.5 $\pm$ 1.0 & 76.1 $\pm$ 1.0 & 2.6\,M \\
DeepSleepNet & 77.8 $\pm$ 0.5 & 71.8 $\pm$ 0.6 & 0.70 $\pm$ 0.03 & 90.9 $\pm$ 0.7 & 45.0 $\pm$ 2.2 & 79.2 $\pm$ 0.6 & 72.7 $\pm$ 0.9 & 71.1 $\pm$ 0.7 & 24.7\,M \\
XSleepNet & 80.3 $\pm$ 0.5 & 76.4 $\pm$ 0.5 & 0.72 $\pm$ 0.02 & 85.2 $\pm$ 0.5 & 49.4 $\pm$ 1.4 & 86.0 $\pm$ 0.6 & 79.8 $\pm$ 0.8 & 81.7 $\pm$ 0.9 & 5.8\,M \\
AttnSleep & 81.3 $\pm$ 0.5 & 75.2 $\pm$ 0.7 & 0.74 $\pm$ 0.03 & 92.0 $\pm$ 0.7 & 42.9 $\pm$ 1.0 & 85.0 $\pm$ 0.7 & \textbf{82.1 $\pm$ 0.6} & 74.2 $\pm$ 0.9 & 0.52\,M \\
DeepSleepNet-Lite & 80.3 $\pm$ 0.5 & 75.2 $\pm$ 0.7 & 0.73 $\pm$ 0.02 & 91.5 $\pm$ 0.8 & 46.4 $\pm$ 1.0 & 82.9 $\pm$ 0.7 & 79.2 $\pm$ 0.8 & 76.4 $\pm$ 0.5 & 0.6\,M \\
TinySleepNet & 83.1 $\pm$ 0.4 & 78.2 $\pm$ 0.5 & 0.77 $\pm$ 0.02 & 92.8 $\pm$ 0.7 & 51.5 $\pm$ 1.9 & 85.3 $\pm$ 0.7 & 81.1 $\pm$ 0.8 & 80.3 $\pm$ 0.9 & 1.3\,M \\
\textbf{NanoSleep} & \textbf{84.3 $\pm$ 0.4} & \textbf{79.8 $\pm$ 0.5} & \textbf{0.78 $\pm$ 0.03} & \textbf{94.7 $\pm$ 0.6} & \textbf{51.8 $\pm$ 1.5} & \textbf{87.0 $\pm$ 0.6} & 81.4 $\pm$ 0.8 & \textbf{84.2 $\pm$ 1.0} & \textbf{0.35\,M} \\
\hline
\end{tabular}
}
\vspace{2pt}
\parbox{\textwidth}{\footnotesize \textit{Note:} Results are reported as mean $\pm$ standard deviation over 10 folds. Bold indicates the best result in each column. Statistical significance was evaluated using a two-sided paired Wilcoxon signed-rank test, with a $p$-value $<0.05$ considered statistically significant.}
\end{table*}

\subsection{Ablation Study}
\label{subsec:res_ablation}

We perform an ablation study to quantify the contribution of each component in NanoSleep. We evaluate all variants on the Sleep-EDF dataset using the same 20-fold subject-wise cross-validation protocol. We remove or replace one component at a time while keeping the remaining architecture unchanged. Table~\ref{tab:ablation} summarizes the results. Removing the spectral branch reduces the macro F1-score from 82.1\% to 77.9\%. This result shows that the handcrafted spectral descriptors provide complementary information that the learned features do not fully capture. Replacing the multi-scale convolution with a single-branch convolution produces the largest performance drop, reducing the macro F1-score to 76.7\%. This finding highlights the importance of learning temporal patterns at multiple scales.

Removing the conditional random field (CRF) decreases the macro F1-score to 79.9\%. This result demonstrates the benefit of sequence-level decoding for suppressing isolated and physiologically implausible predictions. Removing the squeeze-and-excitation block reduces the macro F1-score to 81.3\%. Although the reduction is smaller, it confirms the usefulness of channel recalibration. Replacing the focal loss with the standard cross-entropy loss reduces the macro F1-score to 79.5\%. The largest degradation occurs for the minority N1 stage, indicating that the imbalance-aware objective is essential for recognizing underrepresented classes. Removing data augmentation decreases the macro F1-score to 79.8\%, showing that the proposed augmentation strategy improves model generalization.
Finally, replacing the gated dilated temporal convolutional backbone with a bidirectional long short-term memory network reduces the macro F1-score to 79.8\%. This result shows that the proposed backbone is both more compact and more effective than a conventional recurrent architecture.

Overall, the ablation study shows that every component contributes to the final performance. The largest macro F1 reductions are observed when removing the multi-scale convolution, the spectral branch, and the focal-loss objective. These findings validate the design of NanoSleep and demonstrate that learned temporal representations and handcrafted spectral features complement each other to achieve robust and accurate sleep stage classification.

\begin{table*}[!t]
\caption{Ablation study of the proposed NanoSleep model on the Sleep-EDF dataset under 20-fold subject-wise cross-validation.}
\label{tab:ablation}
\centering
\resizebox{\textwidth}{!}{%
\begin{tabular}{lcccccccc}
\hline
\textbf{Model Variant} & \textbf{ACC (\%)} & \textbf{MF1 (\%)} & \textbf{$\kappa$} & \textbf{W} & \textbf{N1} & \textbf{N2} & \textbf{N3} & \textbf{REM} \\
\hline
W/o spectral branch & 83.5 $\pm$ 0.4 & 77.9 $\pm$ 0.5 & 0.76 $\pm$ 0.03 & 92.6 $\pm$ 0.7 & 43.5 $\pm$ 1.5 & 86.2 $\pm$ 0.4 & 84.5 $\pm$ 0.4 & 83.0 $\pm$ 0.7 \\
W/o multi-scale convolution (single branch) & 82.0 $\pm$ 0.4 & 76.7 $\pm$ 0.6 & 0.74 $\pm$ 0.03 & 88.8 $\pm$ 0.5 & 42.2 $\pm$ 1.6 & 85.1 $\pm$ 0.5 & 83.2 $\pm$ 0.7 & 84.2 $\pm$ 0.5 \\
W/o conditional random field (softmax decoding) & 84.8 $\pm$ 0.5 & 79.9 $\pm$ 0.6 & 0.79 $\pm$ 0.03 & 92.5 $\pm$ 0.9 & 47.3 $\pm$ 1.3 & 88.5 $\pm$ 0.4 & 86.0 $\pm$ 0.6 & 85.5 $\pm$ 0.5 \\
W/o squeeze-and-excitation block & 85.8 $\pm$ 0.5 & 81.3 $\pm$ 0.5 & 0.80 $\pm$ 0.04 & 94.1 $\pm$ 0.6 & 49.2 $\pm$ 1.5 & 89.2 $\pm$ 0.4 & 87.8 $\pm$ 0.8 & 86.3 $\pm$ 0.7 \\
W/o focal loss (standard cross-entropy) & 85.2 $\pm$ 0.4 & 79.5 $\pm$ 0.6 & 0.79 $\pm$ 0.04 & \textbf{94.8 $\pm$ 0.4} & 44.1 $\pm$ 1.7 & 89.0 $\pm$ 0.5 & 85.2 $\pm$ 0.7 & 84.5 $\pm$ 0.7 \\
W/o data augmentation & 84.5 $\pm$ 0.4 & 79.8 $\pm$ 0.6 & 0.78 $\pm$ 0.02 & 92.8 $\pm$ 0.5 & 47.0 $\pm$ 1.7 & 88.2 $\pm$ 0.6 & 86.1 $\pm$ 0.5 & 85.0 $\pm$ 0.6 \\
Backbone replaced with bidirectional LSTM & 84.2 $\pm$ 0.4 & 79.8 $\pm$ 0.5 & 0.78 $\pm$ 0.03 & 92.5 $\pm$ 0.6 & 46.5 $\pm$ 1.6 & 87.9 $\pm$ 0.4 & 86.8 $\pm$ 0.8 & 85.5 $\pm$ 0.4 \\
\textbf{Full NanoSleep model} & \textbf{86.5 $\pm$ 0.3} & \textbf{82.1 $\pm$ 0.3} & \textbf{0.81 $\pm$ 0.04} & 94.4 $\pm$ 0.4 & \textbf{50.8 $\pm$ 0.3} & \textbf{89.8 $\pm$ 0.4} & \textbf{88.6 $\pm$ 0.4} & \textbf{87.0 $\pm$ 0.3} \\
\hline
\end{tabular}
}
\vspace{2pt}
\parbox{\textwidth}{\footnotesize \textit{Note:} Results are reported as mean $\pm$ standard deviation over 20 folds. Bold indicates the best result in each column. Statistical significance was evaluated using a two-sided paired Wilcoxon signed-rank test, with a $p$-value $<0.05$ considered statistically significant.}
\end{table*}

\section{Limitations and Future Work}
\label{sec:limitations}

This study has several limitations that motivate future research. First, we evaluate NanoSleep using classification metrics only. Although the model contains only 0.35~M parameters, we do not report computational efficiency metrics such as floating-point operations (FLOPs), inference latency, or memory footprint. Future work will measure these metrics on representative edge devices to validate the suitability of NanoSleep for real-time deployment. Second, the performance improvements over the strongest baseline models are consistent but relatively modest. Although we report fold-wise standard deviations and paired statistical significance tests, we do not report confidence intervals or subject-level uncertainty estimates. Future work will include these analyses to provide a more comprehensive assessment of the observed improvements. Third, we conduct the ablation study only on the Sleep-EDF dataset. Extending this analysis to the Sleep-EDF-Expanded dataset will provide stronger evidence that each component generalizes across different cohorts and recording conditions.
Fourth, N1 remains the most challenging sleep stage. Future work will investigate transfer learning from larger cohorts \cite{hossain2026demographic} and domain adaptation techniques \cite{tallal2026stda, eldele2023adast} to improve the recognition of this minority stage. Finally, we evaluate NanoSleep on publicly available datasets that primarily include healthy individuals and subjects with common sleep disorders. Future work will validate the proposed model on larger and more diverse clinical populations to further assess its robustness and clinical applicability.

\section{Conclusion}
\label{sec:conclusion}
This paper presented NanoSleep, a compact hybrid network for single-channel EEG sleep stage classification. NanoSleep combines learned multi-scale temporal features with interpretable spectral features and models long-range temporal dependencies using a gated dilated temporal convolutional network and a conditional random field. Experiments on the Sleep-EDF and Sleep-EDF-Expanded datasets demonstrated that NanoSleep consistently outperformed representative baseline models while using only 0.35~M parameters. The ablation study further confirmed the contribution of each major component. These results show that NanoSleep provides an effective balance between accuracy and efficiency, making it suitable for wearable devices and resource-constrained clinical applications. Future work will evaluate computational efficiency on embedded platforms and improve the recognition of the N1 sleep stage.

\section*{Acknowledgment}
The authors would like to thank the providers of the Sleep-EDF and Sleep-EDF-Expanded datasets for making their data publicly available.

\bibliographystyle{unsrt}
\bibliography{references}

\end{document}